\documentclass{sigkddExp}

\usepackage{bold-extra}
\usepackage{amsmath}
\usepackage{mathtools}
\usepackage{multirow}
\usepackage{xcolor}
\usepackage{makecell}
\usepackage{graphics}
\usepackage{adjustbox}
\usepackage{enumitem,booktabs,cfr-lm}
\usepackage[referable]{threeparttablex}
\usepackage{tabularx}
\usepackage{array}
\usepackage{amssymb}
\usepackage{url}
\usepackage{diagbox}
\renewlist{tablenotes}{enumerate}{1}
\makeatletter
\setlist[tablenotes]{label=\tnote{\alph*},ref=\alph*,itemsep=\z@,topsep=\z@skip,partopsep=\z@skip,parsep=\z@,itemindent=\z@,labelindent=\tabcolsep,labelsep=.2em,leftmargin=*,align=left,before={\footnotesize}}
\makeatother

\usepackage{hyperref}       % hyperlinks            
\hypersetup{
    colorlinks=true,
    citecolor=red,
    urlcolor=red,
}
\usepackage{inconsolata}
\usepackage{amsmath,amsfonts}
\usepackage{array}
\usepackage[caption=false,font=normalsize,labelfont=sf,textfont=sf]{subfig}
\usepackage{textcomp}
\usepackage{stfloats}
\usepackage{url}
\usepackage{verbatim}
\usepackage{graphicx}
\usepackage{cite}
\usepackage{subfloat}
\usepackage{url}
\usepackage{graphicx}
\usepackage{csquotes}
\usepackage{graphicx}
\usepackage{amsmath}
\usepackage{mathtools}
\usepackage{amssymb}
\usepackage{amsfonts}
\usepackage{bm}
\usepackage{booktabs}
\usepackage{bbm}
\usepackage{multirow}
\usepackage{enumitem}
\usepackage{xcolor}
\usepackage{rotating}
\usepackage{makecell}
\usepackage[capitalize,noabbrev,nameinlink]{cleveref}
\usepackage{listings}
\usepackage{tabularx}
\usepackage{colortbl}
\usepackage{xspace}
\newcommand{\yarrow}{\vcenter{\hbox{\includegraphics[scale=0.2]{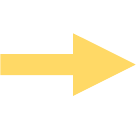}}}}

\definecolor{red}{RGB}{104,52,154}

\setenumerate[1]{itemsep=0pt,partopsep=0pt,parsep=\parskip,topsep=5pt}
\setitemize[1]{itemsep=0pt,partopsep=0pt,parsep=\parskip,topsep=5pt}
\setdescription{itemsep=0pt,partopsep=0pt,parsep=\parskip,topsep=5pt}

\newcommand{\Cora}{\texttt{Cora}\xspace}
\newcommand{\CiteSeer}{\texttt{CiteSeer}\xspace}
\newcommand{\PubMed}{\texttt{PubMed}\xspace}
\newcommand{\CoraFull}{\texttt{CoraFull}\xspace}
\newcommand{\Photo}{\texttt{Photo}\xspace}
\newcommand{\Physics}{\texttt{Physics}\xspace}
\newcommand{\SFairGNN}{\texttt{SFairGNN}\xspace}

\begin{document}

\title{Marginal Nodes Matter: Towards Structure Fairness \\ in Graphs}

\author{Xiaotian Han$^1$  Kaixiong Zhou$^2$  Ting-Hsiang Wang$^1$  Jundong Li$^3$  Fei Wang$^4$ Na Zou$^1$ \\
\texttt{$^1$Texas A\&M University, $^2$Rice University, $^3$University of Virginia, $^4$Weill Cornell Medicine}\\
\texttt{\{han,thwang1231,nzou1\}@tamu.edu, jundong@virginia.edu, few2001@med.cornell.edu}
}

% \date{24 June 2023}

\maketitle
\begin{abstract}
In social network, a person located at the periphery region (marginal node) is likely to be treated unfairly when compared with the persons at the center. While existing fairness works on graphs mainly focus on protecting sensitive attributes (e.g., age and gender), the fairness incurred by the graph structure should also be given attention. On the other hand, the information aggregation mechanism of graph neural networks amplifies such structure unfairness, as marginal nodes are often far away from other nodes. In this paper, we focus on novel fairness incurred by the graph structure on graph neural networks, named \emph{structure fairness}. Specifically, we first analyzed multiple graphs and observed that marginal nodes in graphs have a worse performance of downstream tasks than others in graph neural networks. Motivated by the observation, we propose \textbf{S}tructural \textbf{Fair} \textbf{G}raph \textbf{N}eural \textbf{N}etwork (\SFairGNN), which combines neighborhood expansion based structure debiasing with hop-aware attentive information aggregation to achieve structure fairness. Our experiments show \SFairGNN can significantly improve structure fairness while maintaining overall performance in the downstream tasks.
\end{abstract}

\section{Introduction}\label{sec:Introduction}
Recent years have witnessed a surge of research interests in graph machine learning for different applications, such as social networks~\cite{fan2019graph,tan2019deep}, protein-protein interaction networks~\cite{airola2008all}, knowledge graphs~\cite{zhang2020relational,nguyen2022node,dong2023hierarchy}, and Spatio-Temporal forecasting~\cite{9780945}. Among them, graph neural networks (GNNs)

~\cite{wu2020comprehensive,zhang2020deep,zhou2020graph,wu2022graph}, as a family of deep learning based algorithms, have become an essential paradigm for graph analysis due to their superior modeling performance. Despite their success, it has been shown that these algorithms often suffer from fairness issues for decision-making~\cite{du2020fairness,rahman2019fairwalk,zhang2022fairness,choudhary2022survey,dong2022fairness,ma2022learning,dai2022comprehensive,wang2022improving}. To mitigate this issue, a vast majority of existing works on graph fairness~\cite{bose2019compositional,du2020fairness,rahman2019fairwalk,ma2022learning,song2022guide} focus on obtaining fair node representations that are invariant to the change of the protected attributes such as age, gender, and ethnicity.

Complementing existing efforts on learning fair representations regarding protected attributes~\cite{bose2019compositional,rahman2019fairwalk,du2020fairness,song2022guide,ma2022learning}, we investigate a novel problem on the fairness issue of GNNs from the \emph{structure} perspective. In particular, graph structure fairness should be paid more attention to because of the following key reasons: 
1) Marginal nodes in the graph should not be automatically depreciated by graph learning algorithms. For instance, in the Twitter network, a tweet posted by a user with few followers may receive limited attention, although the tweet could be insightful and valuable for the whole community. As another example, in a job market network such as the LinkedIn network, people should not be neglected just because they are at the periphery of the social network, a situation commonly faced by individuals from underprivileged communities.
2) As one of the most popular methods in graph analysis, GNNs exacerbate structure unfairness because their built-in information aggregation mechanism exploits the egocentric structure of nodes. In other words, each node only receives information from nearby nodes in information aggregation.

Despite its importance and necessity, understanding and achieving structure fairness on GNNs is non-trivial due to the following two challenges. First, a proper measurement of structure fairness is missing, although a quantitative measurement is essential to detect and compensate for \emph{structurally} underprivileged nodes. Second, given that GNNs use an information aggregation mechanism to update node representation from neighbors, but the information aggregation mechanism implicitly causes unfair decision-making, it remains a challenging problem to design an effective graph neural network framework to mitigate the unfairness problem of information aggregation.

To overcome the challenges mentioned above, we first validate that existing graph neural networks suffer from structure fairness problems and then propose to address them accordingly. We first conduct an experiment to validate that structure fairness is highly correlated to the performance of downstream tasks in existing GNNs. Since the information aggregation mechanism of GNNs essentially enriches the node information by aggregating information from their neighbors, centrality~\cite{borgatti2005centrality,marsden2002egocentric} is an intuitive indicator of graph structure, which contains the information between nodes and their neighbors. To this end, we adopt two centrality measurements~(closeness centrality~\cite{koschutzki2005centrality} and eigenvector centrality~\cite{bonacich2007some}) to identify the marginal nodes.

To address structure fairness in graph neural networks, we propose a simple-yet-effective \textbf{S}tructural \textbf{Fair} \textbf{G}raph \textbf{N}eural \textbf{N}etwork (\SFairGNN), which mitigates the structure unfairness issue of GNNs without sacrificing their performance on downstream tasks. Specifically, \SFairGNN has two essential components: 1) neighborhood expansion and 2) hop-aware attentive information aggregation. Neighborhood expansion compensates marginal nodes by introducing new neighbors to them during the training phase. On the other hand, hop-aware attentive information aggregation optimizes how the new information should be properly and effectively integrated into each node based on downstream tasks. Our investigation and proposed model will help researchers and society understand how to identify and mitigate structure unfairness in graph-based decision-making. In summary, the \textbf{contributions} of this work are as follows:

\begin{itemize}[leftmargin=0.4cm]
    \item We conducted an analysis of the performance of downstream tasks on marginal nodes and found that they were significantly disadvantaged, leading to unfair decision-making. Our investigation led us to the initial observation that the graph structure was responsible for the unfairness suffered by these nodes.
    \item To address the issue of unfairness caused by the graph structure, we proposed a new graph neural network model called \SFairGNN. Our proposed model is specifically designed to mitigate the impact of structural unfairness in graph neural networks.
    \item To evaluate the performance of \SFairGNN, we conducted experiments on real-world datasets. Our results showed that \SFairGNN outperformed state-of-the-art GNN models in terms of fairness, without sacrificing performance. Our experiments demonstrate the effectiveness of \SFairGNN in reducing unfairness and improving overall performance in graph-based machine learning models.
\end{itemize}

The rest of the paper is organized as follows. In \cref{sec:Problem Formulation}, we provide an overview of the necessary background for \SFairGNN, which includes an introduction to graph neural networks and metrics for measuring the structure of graphs. \cref{sec:Problem Formulation} delves into the issue of fairness in graph structures. Next, in \cref{sec:method}, we present our proposed fairness-aware graph neural network, \SFairGNN. We discuss the experimental findings and analysis in \cref{sec:exp}. Finally, in \cref{sec:Related Works,sec:Conclusion and Future work}, we discuss the related work and provide concluding remarks and suggestions for future research.

\section{Preliminaries}\label{sec:Preliminary}

In this section, we introduce the essential preliminaries for our proposed methods, including graph neural networks, as well as two centrality metrics of a graph: closeness centrality and eigenvector centrality.

\subsection{Graph Neural Networks} Most graph neural networks~(GNNs)~\cite{hamilton2017inductive,velivckovic2017graph,wu2019simplifying,ling2023graph} adopt a message-passing mechanism to update node representations over the graph iteratively. Without loss of generality, we present a general framework of GNNs, which can be divided into two steps: aggregation and transformation. Generally, at the $k$-th layer in GNNs, the hidden representation $\mathbf{x}^{(k)}_i$ of node $n_i$ is updated as follows:
\begin{equation*}\label{eq:GNN}
    \begin{split}
        \mathbf{h}^{(k)}_i & = \mathrm{AGGREGATE}(\{a^{(k)}_{ij} \mathbf{W}^{(k)} \mathbf{x}^{(k-1)}_j: j\in \mathcal{N}_i\}), \\
        \mathbf{x}^{(k)}_i & = \mathrm{ACT}(\mathrm{COMBINE}(\mathbf{x}^{(k-1)}_i, \mathbf{h}^{(k)}_i)),
    \end{split}
\end{equation*}
\noindent where $\mathbf{W}^{(k)}$ denotes the trainable weight, and $a_{ij}^{(k)}$ denotes the edge weight between nodes $i$ and $j$, and $\mathcal{N}_i$ denotes the set of neighbors adjacent to node $i$. $\mathrm{AGGREGATE}(\cdot)$ is an aggregation function, which is applied to aggregate the node representations of all neighbors. Finally, $\mathrm{COMBINE}(\cdot)$ and $\mathrm{ACT}(\cdot)$ are applied to combine neighbor information and  activate node representations, respectively.

\subsection{Indicators of Structure}
To account for the impact of graph structure toward fair decision-making in GNNs, we first need to identify the marginal nodes in the graph. To reflect node sociological origin~\cite{koschutzki2005centrality} and the graph structure at the node level, we use two important centralities~(i.e., closeness centrality and eigenvector centrality) to quantify the structure of each node. Closeness centrality~\cite{koschutzki2005centrality} of a node measures its average nearness~(inverse distance) to all other nodes in the graph. Eigenvector centrality~\cite{bonacich2007some} of a node measures the influence it has on all other nodes in the graph.

\textbf{Closeness Centrality}
Closeness centrality~\cite{bavelas1950communication,sabidussi1966centrality} of a node is the reciprocal of the sum of the shortest path distances from the node to all other nodes on the graph. The nodes with a high closeness centrality score have the shortest distances to all other nodes, which easily access other nodes and influence other nodes. For instance, in a social network, a person with a lower mean distance from others might find that their opinions reach others more quickly than the opinions of someone with a higher mean distance. Formally, the mathematical expression of closeness centrality of node $n_i$ is:
\begin{equation}\label{equ:clossesness}
    c(n_i) = \frac{N-1}{\Sigma_{n_i}d(n_i,n_j)},
\end{equation}
where $d(n_i,n_j)$ is the shortest-path distance between nodes $i$ and $j$, $N$ is the number of nodes in the graph.

\textbf{Eigenvector Centrality}
Eigenvector centrality is a measure of the influence of each node in a graph~\cite{bonacich2007some}. A high eigenvector centrality score for a node indicates that it is connected to many other nodes with high scores. The eigenvector centrality score for node $i$ is calculated as the $i$-th element of the vector $\mathbf{x}$, which is defined by the equation $\mathbf{Ax} = \lambda \mathbf{x}$, where $\mathbf{A}$ is the adjacency matrix of the graph and $\lambda$ is the corresponding eigenvalue. According to the Perron-Frobenius theorem~\cite{pillai2005perron}, if $\lambda$ is the largest eigenvalue of $\mathbf{A}$, then there exists a unique solution $\mathbf{x}$, all of whose entries are positive.\cite{hagberg2008exploring} Throughout this paper, we use the indicator variable $s_i$ to represent the centrality metric of node $i$. Specifically, $s_i$ can take either closeness centrality or eigenvector centrality, depending on the context.

\section{Problem Formulation}\label{sec:Problem Formulation}
In the following subsections, we introduce and analyze the structure fairness in graph neural networks and propose how to measure the structure fairness. 

\begin{figure*}[t]
    \centering
    \includegraphics[width=\textwidth]{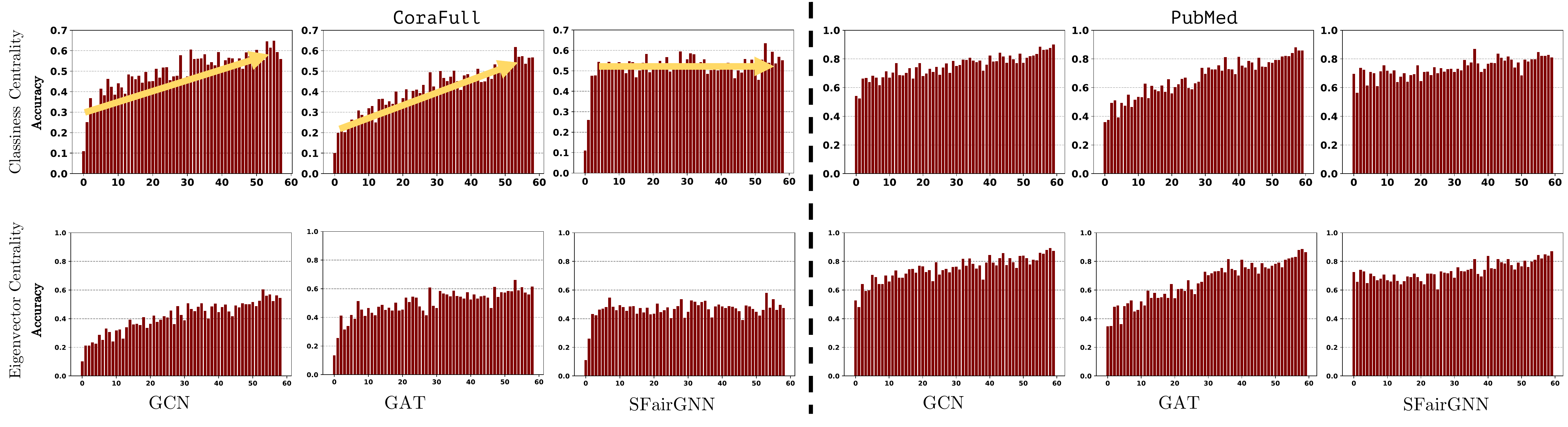}
    \vspace{-20pt}
    \caption{The distribution of node classification accuracy with respect to closeness/eigenvector centrality on \CoraFull and \PubMed datasets. The X-axis indicates closeness/eigenvector centrality and Y-axis indicates classification accuracy. We can see that the performance of GCN and GAT is biased towards nodes with higher closeness/eigenvector centrality. In contrast, \SFairGNN has a more balanced classification performance distribution and implies a lower correlation between downstream task performance and closeness/eigenvector centrality, implying a lower correlation between downstream task performance and centrality scores. $\yarrow$ on the top-left figure clearly indicates that GCN and GAT are biased towards nodes with higher closeness/eigenvector centrality, while our method does not exhibit this bias.}\label{fig:figure1}
\end{figure*}

\subsection{Preliminary Experiments}\label{sec:pf:fairness_exp}
To explore the structure unfairness in GNNs, we perform node classification task with representative GNNs, i.e., graph convolutional networks (GCN)~\cite{kipf2016semi} and graph attention networks (GAT)~\cite{velivckovic2017graph}. Specially, we cluster nodes in the graph into multiple bins based on their indicators of structure (i.e., closeness centrality and eigenvector centrality). Then we calculate the average accuracy of each bin and compare the accuracy difference between bins.

\textbf{Experimental Results.} We present the experimental results on \CoraFull and \PubMed dataset with closeness and eigenvector centrality in \cref{fig:figure1}. The X-axis represents different levels of closeness/eigenvector centrality, while Y-axis represents the prediction accuracy.  

\textbf{Results Analysis.} The lower classification accuracy of marginal nodes shows that they are treated unfairly by GCN and GAT. Therefore, it is clear that structure unfairness in GCN and GAT can lead to discriminative results for marginal individuals, especially when life-changing decisions are at stake. Such unfairness can lead to discriminatory results for marginalized individuals, particularly in cases where significant decisions are at stake. 
As shown in the figures on the top row, GCN and GAT tend to achieve higher classification accuracy for nodes with higher closeness centrality and lower classification accuracy for nodes with lower closeness centrality. The experimental results presented in this section indicate that \SFairGNN is capable of achieving structural fairness for downstream tasks.

Similarly, the experiment results on another indicator of structure~(i.e., eigenvector centrality) have a similar trend. From the figure, GCN and GAT bias towards nodes with higher eigenvector centrality. In contrast, \SFairGNN has comparable average node classification accuracy and has more even performance distribution and a lower correlation between performance and eigenvector centrality.  The figures in the second row in \cref{fig:figure1} show that eigenvector centrality is indeed highly correlated to classification accuracy with GCN and GAT models. The observation here is quite similar to the experiment using closeness centrality as an indicator of structure.

\subsection{Why GNNs are Structurally Unfair?}
Here, we analyze the root cause of the unfair decision-making of GNNs. 
As we have discussed, the neighborhood aggregation mechanism of GNNs treats nodes with different structures unfairly, especially for the marginal nodes. To further illustrate the unfair neighbor aggregation mechanism, we use \cref{fig:model}a as a toy example. There are three groups of nodes with different egocentric structures, which are marked in yellow, blue, and green. The yellow indicates the central nodes, while the green indicates the marginal nodes. And the blue nodes situate between them. Since the green nodes and the blue nodes obtain less information than the yellow nodes, the GNNs will treat these three groups of nodes differently. Through the analysis of the toy example, we conclude that the aggregation mechanism of GNNs can lead to structure unfairness and put the marginal nodes into a disadvantaged situation since it is inherently difficult for marginal nodes to receive the global information of the graph. Although the aggregation mechanism in GNNs improves the performance of the graph analysis, it amplifies the structure fairness.

\subsection{Measurement of Structure Fairness}
Since centralities and classification accuracy are both continuous variables, we propose the following two metrics to measure structure fairness. One is the correlation between the indicator of graph structure and the probability of correct classification. Another one is the variation of the average classification accuracy for nodes in different bins within different egocentric structure ranges.

\vspace{5pt}
\noindent\textbf{Pearson Correlation Coefficient (PCC)} is a statistical metric that measures the linear correlation between two variables~\cite{benesty2009pearson}. In our scenario, we use PCC to assess the linear correlation between closeness centrality and the probability of correct classification. A PCC value in the range of $[-1,0]/[0,1]$ indicates that two variables are negatively/positively correlated, respectively. A low PCC value suggests that the algorithm is preserving structural fairness.

\vspace{5pt}
\noindent\textbf{Standard Deviation~(STD)} quantifies the variation of average classification accuracy of nodes in different bins within different ranges of centrality values. Specifically, we cluster nodes into different bins based on their centrality values. We first calculate the mean classification accuracy for each bin and then calculate their STD. A small STD value means that a model has similar classification accuracy for nodes in different bins, which implies that nodes are treated fairly. From \cref{fig:figure1}, the STD of the GCN, GAT, and \SFairGNN are $0.2480, 0.2935, 0.1583$ for the \CoraFull dataset, respectively. As our proposed \SFairGNN has the lowest STD, it beats GNN and GAT in terms of structure fairness based on the closeness/eigenvector centrality.

\section{The Proposed Method}\label{sec:method}
\begin{figure}[t]
    \centering
    \vspace{20pt}
    \includegraphics[width=0.47\textwidth]{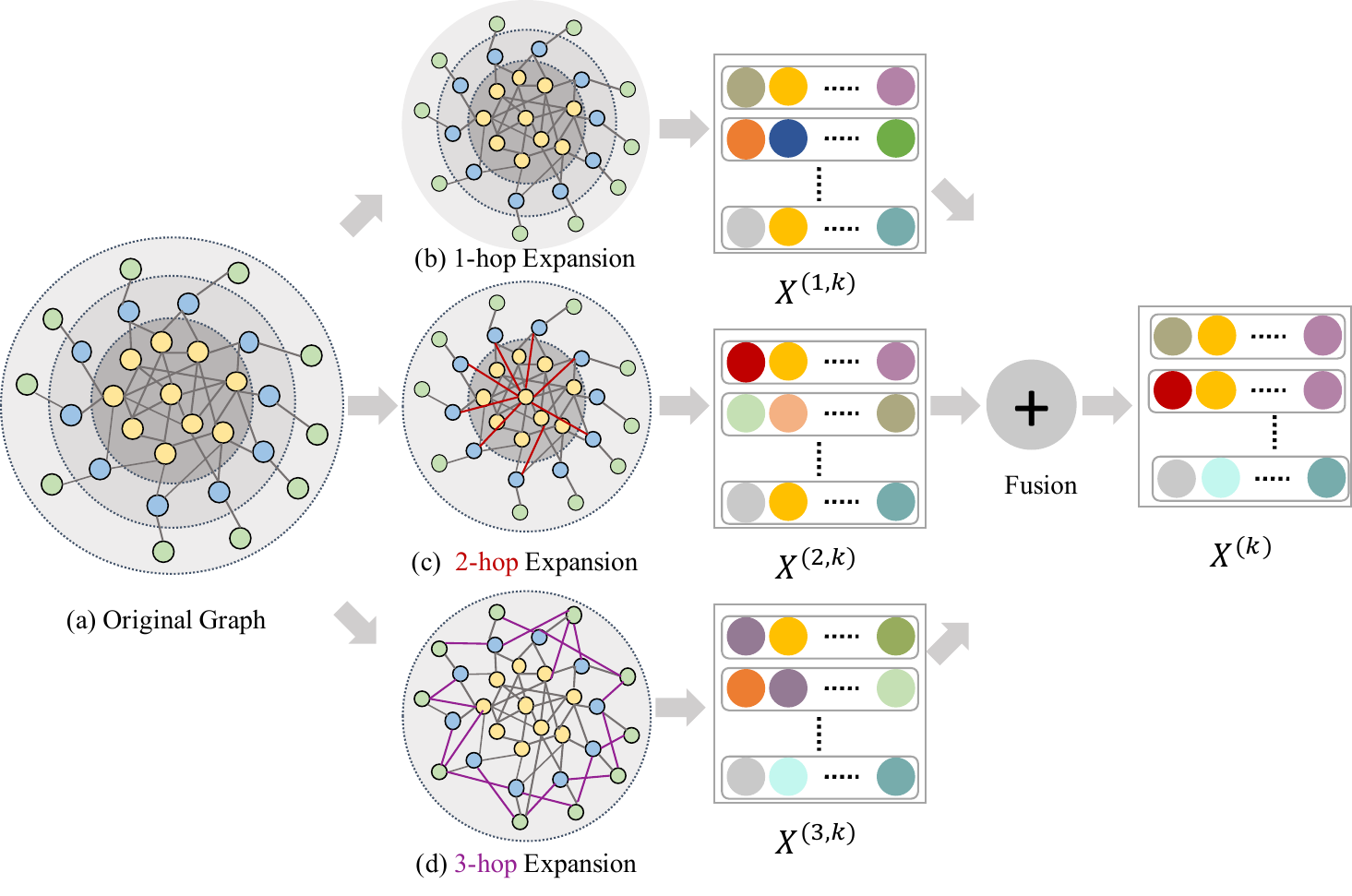}
    \vspace{-5pt}
     \caption{An illustration of our proposed method, specifically an example of the $2$-layer \SFairGNN. After the neighborhood expansion and hop-wise information aggregation, we obtain the embedding matrices for these three hops: $\{\mathbf{X}^{(1,k)}, \mathbf{X}^{(2,k)},\mathbf{X}^{(3,k)}\}$. They are combined with a fusion function to output the final embedding $\mathbf{X}^{(k)}$ at $k$-th graph convolution layer.}\label{fig:model} 
\end{figure}

In this section, we introduce \textbf{S}tructural \textbf{Fair} \textbf{G}raph \textbf{N}eural \textbf{N}etwork (\SFairGNN), whose fundamental intuition is to allow marginal nodes to better obtain the information from the center of the graph. The framework of \SFairGNN shown in \cref{fig:model} is composed of two essential components: 1) the neighborhood expansion component that creates new neighbors for marginal nodes in the graph and brings them closer to the center of the graph; 2) the hop-aware attentive information aggregation component that leverages attention mechanism to fuse representations of neighbors at different hops towards a fair embedding for all nodes. In the following sections, we will introduce the details of neighborhood expansion and hop-aware attentive information aggregation. To illustrate the proposed \SFairGNN, we use closeness centrality as an example indicator of graph structure.

\begin{figure}[!t]
\centering
\subfloat[Original]{\includegraphics[height=1.0in]{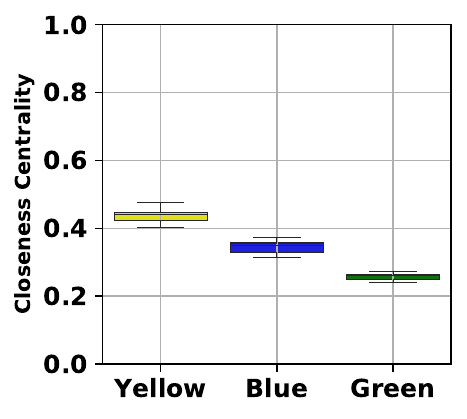}\label{fig:closeness:original}}
\subfloat[2-hop]{\includegraphics[height=1.0in]{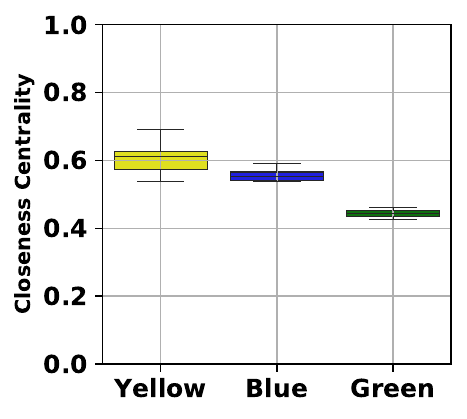}\label{fig:closeness:graph_expansion1}}
\subfloat[3-hop]{\includegraphics[height=1.0in]{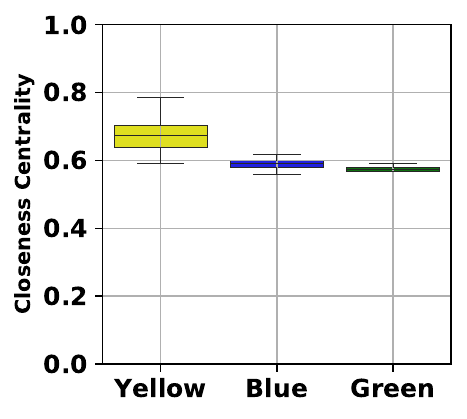}\label{fig:closeness:graph_expansion2}}
  \caption{The distribution of closeness centrality for different-hop neighbors. The X-axis represents nodes colored yellow, blue, and green, while the Y-axis denotes closeness centrality. In (a), we have the original graph, while (b) and (c) show the graphs after $2$-hop and $3$-hop neighborhood expansions, respectively. We observe that the distribution of closeness centrality becomes more stable as the hop number $h$ increases.}\label{fig:closeness}
\end{figure}

\subsection{Neighborhood Expansion Based Structure Debiasing}\label{sec:method:Neighbor Expand}

Before delving into the neighborhood expansion based structure debiasing, we first investigate the distribution of closeness centrality over a synthetic graph to better understand the unfairness of the traditional information aggregation of GNNs. We plot the distribution of closeness centrality in \cref{fig:closeness:original} based on the graph shown in \cref{fig:model}a. The synthetic graph in \cref{fig:model}a contains central, middle, and marginal nodes colored yellow, blue, and green. It is observed from \cref{fig:closeness:original} that the closeness centrality varies significantly among the three colors, where the marginal nodes are much lower than the central nodes.

Motivated by the above observation, we propose the neighborhood expansion method to debias the structure unfairness by strengthening the marginal nodes' connections to the center of the graph. Specifically, a heuristic threshold is adopted to determine whether a node is marginal or not. We name this threshold as {margin line} and denote it as $line$. The  margin line serves as an important hyperparameter of the proposed method.  The nodes with closeness centrality $s_{i} \leq line$ are regarded as marginal nodes; otherwise, they are central nodes. For each node $n_i$, we expand its original neighbor set $\mathcal{N}_i$ to a hyperset $\{ \mathcal{N}^{(1)}_i, \mathcal{N}^{(2)}_i, \cdots, \mathcal{N}^{(h)}_i \}$ to indicate neighbors within $h$ hops. The $1$-hop neighbor set $\mathcal{N}_i^{(1)}$ is given by $\mathcal{N}_i$, which is generated based on the original adjacency matrix $\mathbf{A}$. We define a debiased adjacency matrix $\tilde{\mathbf{A}}$ with each element given as:
\begin{equation}
	\tilde{\mathbf{A}}_{i,j}= 
	\begin{cases}
	\mathbf{A}_{i,j}, &\text{if}~~s_{i} <= line~ \text{or}~s_{j} <= line,\\
	0, &~~\text{otherwise},
    \end{cases}
\end{equation}
where $s_i$ is the indicator of structure for node $n_i$. In the debiased matrix $\tilde{\mathbf{A}}$, we only strengthen the connections of marginal nodes. Then we use the $h$-order power of debiased matrix $\tilde{\mathbf{A}}^h$ to generate neighbor set $\mathcal{N}_i^{(h)}$. We obtain a series of neighbor sets of node $n_i$ from different hops $\{ \mathcal{N}^{(1)}_i, \mathcal{N}^{(2)}_i, \cdots, \mathcal{N}^{(h)}_i \}$, which will be used for hop-aware attentive information aggregation.

\cref{fig:model}b, \cref{fig:model}c, and \cref{fig:model}d represent the graphs obtained by neighborhood expansions with $h = 1, 2$ and $3$, respectively. Note that the graph with $1$-hop expansion is the same as the original graph. To verify the effectiveness of neighborhood expansion, we illustrate the closeness centrality distributions of graphs with different hop expansions in \cref{fig:closeness}. As we can see, increasing hop numbers for neighborhood expansion shrinks the closeness centrality gap between different node groups. The empirical study shows that neighborhood expansion can effectively alleviate the structure unfairness in terms of closeness centrality.

\subsection{Hop-Aware Attentive Aggregation}\label{sec:method:Information Aggregation}
Since the neighbors of different hops contain different amounts of information, we propose hop-aware attentive information aggregation to provide customized information aggregation from neighbors of different hops. The overview is illustrated in \cref{fig:model}. Specifically, at the $k$-th graph convolution layer, let $\mathbf{X}^{(k-1)}$ and $\mathbf{X}^{(k)}$ denote the input and output embedding matrices of all the nodes, respectively. The signal processing at each layer includes two steps: 1) hop-wise information aggregation and 2) hop fusion. With the series of neighbor sets $\{ \mathcal{N}^{(1)}_i, \mathcal{N}^{(2)}_i, \cdots, \mathcal{N}^{(h)}_i \}$ obtained by the neighborhood expansion, we can compute the hop-wise embedding matrices $\{ \mathbf{X}^{(1,k)}, \mathbf{X}^{(2,k)}, \cdots, \mathbf{X}^{(h,k)} \}$ at the $k$-th layer. Then we fuse these embedding matrices to obtain the final output node representation $\mathbf{X}^{(k)}$. We present the details of the two steps in the following:

\vspace{5pt}
\noindent\textbf{Hop-Wise Information Aggregation}
Given neighbor sets $\mathcal{N}^{(h)}_i$ of all the nodes, we aim to obtain embedding $\mathbf{X}^{(h,k)}$ at the $k$-th layer for $h$-th hop neighbors. Considering the different importance of expanded neighbors, we leverage the attention mechanism~\cite{velivckovic2017graph} to learn the attention coefficients of different hop neighbors automatically. Specifically, given neighbor ${j}$ for node ${i}$, attention coefficients is calculated by:
\begin{equation}\label{equ:attention}
    \alpha_{ij}^{(h,k)} 
    = \frac{\exp\left( f\left(  \mathbf{W}^{(h,k)} \cdot \left[\mathbf{x}_i^{(k-1)}  || \mathbf{x}_j^{(k-1)} \right]  \right) \right)}{\sum_{m \in \mathcal{N}^{(h)}_i}\exp\left( f\left( \mathbf{W}^{(h,k)} \cdot \left[ \mathbf{x}_i^{(k-1)} || \mathbf{x}_m^{(k-1)}  \right] \right) \right) },
\end{equation}
where $||$ is the concatenation operation. $f(\cdot)$ is the activation function. $\mathbf{W}^{(h,k)}$ is the trainable weight matrix for the $h$-th hop neighbors at the $k$-th layer. $\mathbf{x}^{(k-1)}_i$ is the representation of node $i$, which is the $i$-th row of embedding matrix $\mathbf{X}^{(k-1)}$. With the attention coefficient $\alpha_{ij}^{(h,k)}$, the embedding of node $n_i$ at the $k$-th layer within $h$-th hop neighbors is given by:
\begin{equation}
    \label{eq:FairGNN}
        \mathbf{x}^{(h,k)}_i = \mathrm{AGG}\left( \{\alpha^{(h,k)}_{ij} \mathbf{W}^{(h,k)} \mathbf{x}^{(k-1)}_j: j\in \mathcal{N}^{(h)}_i\} \right).
\end{equation}
Then we have $\mathbf{X}^{(h,k)} = [\mathbf{x}^{(h,k)}_1, \mathbf{x}^{(h,k)}_2, \cdots, \mathbf{x}^{(h,k)}_N]^T$, where $N$ is the total number of nodes in the graph.

\vspace{5pt}
\noindent\textbf{Hop Fusion} After we obtain a series of hop-wise embedding matrices $\{ \mathbf{X}^{(1,k)}, \mathbf{X}^{(2,k)}, \cdots, \mathbf{X}^{(h,k)} \}$, we adopt fusion function $\mathcal{F}$ to fuse them into final output embedding: 
\begin{equation}
\label{equ:agge}
\mathbf{X}^{(k)} = \mathcal{F}\left( \mathbf{X}^{(1,k)}, \mathbf{X}^{(2,k)}, \cdots, \mathbf{X}^{(h,k)} \right).
\end{equation}
Fusion methods $\mathcal{F}$ could be implemented by different functions, such as average or maximum functions. We perform experiments to compare the effects of different fusion methods~\cite{scherer2010evaluation} (i.e., SEQ, AVG, and MAX) in \cref{sec:exp:fusion}. Note that in the used datasets, the input node features usually play a deterministic role in deciding the nodes' labels. These informative node features will prohibit us from focusing on the influence of structure on the following classification task. Therefore, instead of taking the node features as input, we randomly initialize $\mathbf{X}^{(0)}$ as input node features and set it trainable during the training phase.

\subsection{Discussion}
\textbf{Model Analysis.} Here we analyze why the two key components of \SFairGNN promote structure fairness: 1) neighborhood expansion strengthens the connection of marginal nodes in the graph, which positions them closer to the center of the graph and thus improves their accessibility to information from other nodes. 2) hop-aware attentive information aggregation enables \SFairGNN to learn customized weights for information from neighbors of different hops away, which adaptively considers the varied strength of influence from neighboring nodes. 

\textbf{Avoid Constraining Loss over Fairness} \SFairGNN promotes the structure fairness of marginal nodes by neighbor expansion instead of explicitly optimizing fairness in loss function for two reasons. First, \SFairGNN intends to improve fairness not for the measure's own sake, but for a “healthier” graph structure.
This minimizes the trade-off between the overall performance and fairness for \SFairGNN, as lowering the performance of ``privileged'' nodes is the shortest path to achieve fairness for direct optimization.
From \cref{tab:Accuracy and standard deviation}, we verify \SFairGNN does not sacrifice the model performance for fairness. Second, compared to direct optimization, neighbor expansion is drastically more efficient. This is because computing centrality scores for every training iteration is prohibitively expensive.

\textbf{Comparison to Related Works.} Our proposed method focuses on structure fairness, while the previous works~\cite{rahman2019fairwalk,dai2020fairgnn,tang2020investigating,bose2019compositional} mainly focus on learning fair node representations regarding protected attributes. For the two methods which resemble \SFairGNN the most, we discuss the difference between them and our proposed method below.
Fairwalk~\cite{rahman2019fairwalk} proposes a fairness-aware embedding method based on the random walk technique for attributed networks. FairGNN~\cite{dai2020fairgnn} uses adversarial debiasing and covariance constraint to regularize the GNN to obtain fair node representations and predictions. Both of them focus on the categorical sensitive attributes, while our proposed method does not require any sensitive features and aims to achieve structure-level debias.

\textbf{Comparison to Cold-Start} 
Our problem is different from the cold-start problem to a large extent, which does not aim to improve fairness across users but improve the recommendation performance for new users. In addition, while cold-start in recommender systems gradually warms up, unfairness in graphs, especially in the social context, can worsen over time and thus an algorithmic solution is needed.
% \textbf{Limitation}

\section{Experiments}\label{sec:exp}
In this section, we empirically evaluate the effectiveness of \SFairGNN in node classification tasks by answering the following questions:

\begin{itemize}[leftmargin=0.4cm]
    \item \textbf{Q1:} To what extent can \SFairGNN outperform the current  GNNs in simultaneously achieving high classification accuracy and maintaining structure fairness?
    \item \textbf{Q3:} What is the impact of using different fusion methods on the performance of \SFairGNN?
    \item \textbf{Q2:} How do the hyperparameters of \SFairGNN, such as the hop number and margin line, affect its ability to mitigate bias?
\end{itemize}

\subsection{Experimental Setting}
In this section, we outlines our experimental setting for evaluating the effectiveness of \SFairGNN in mitigating bias and achieving structural fairness. Specifically, we provide a comprehensive overview of the benchmark datasets, the baseline methods, and the implementation details both our proposed method and the baseline models.

\textbf{Baselines} In the experiments, we consider the widely used  graph neural networks including graph convolutional networks (GCN)~\cite{kipf2016semi}, graph attention networks (GAT)~\cite{velivckovic2017graph}, and GraphSAGE~\cite{hamilton2017inductive} as baseline methods. The details of baseline methods are listed as follows:

\begin{itemize}[leftmargin=0.4cm]
\item \textbf{GCN}~\cite{kipf2016semi} is the pioneer work of graph neural networks. GCN is a scalable approach for semi-supervised learning on graph-structured data that is based on an efficient variant of convolutional neural networks.

\item \textbf{GAT}~\cite{velivckovic2017graph} incorporates the attention mechanism into the propagation step. This method computes the hidden states of each node by attending to its neighbors following a self-attention strategy.

\item \textbf{GraphSAGE}~\cite{hamilton2017inductive} efficiently generates node representations for previously unseen data. Instead of training individual representations for each node, GraphSAGE learns a function that generates representations by sampling and aggregating features from a node’s local neighborhood.
\end{itemize}

\textbf{Datasets} Following the previous works, we use six benchmark datasets to study the graph structure fairness on GNN models. They are \Cora, \CiteSeer, \PubMed, \CoraFull, \Physics, and \Photo~\cite{yang2016revisiting}. The detailed statistics of these datasets are presented in \cref{tab:statistics}.

\begin{table}[t]
\centering
\caption{The statistics of the datasets.}\label{tab:statistics}
\scalebox{0.83}{
  \begin{tabular}{lrrrrrrr}
    \toprule
    \textbf{\textsc{Dataset}}      &\textbf{\#Nodes}   &\textbf{\#Edges} &\textbf{\#Feature}   &\textbf{Density} &\textbf{\#Class} \\
    \midrule
    \Cora                  &$2708 $              &$10556 $       &$1433$         &$0.0014$     &$7 $\\
    \CiteSeer              &$3327 $              &$9104  $       &$3703$         &$0.0008$     &$6 $\\
    \PubMed                &$19717$              &$88648 $       &$500 $         &$0.0002$     &$3 $\\
    \CoraFull              &$19793$              &$130622$       &$8710$         &$0.0003$     &$70$\\
    \Photo                 &$7650 $              &$238162$       &$745 $         &$0.0042$     &$8 $\\
    \Physics               &$34493$              &$495924$       &$8415$         &$0.0004$     &$5 $\\
    \toprule
  \end{tabular}
}
\end{table}

\begin{itemize}[leftmargin=0.4cm]
\item \textbf{\Cora}, \textbf{\CiteSeer}, and \textbf{\PubMed} are citation networks~\cite{yang2016revisiting}, where nodes and edges denote papers and citations, respectively. Node features are bay-of-words for papers and node labels indicate the fields of papers.

\item \textbf{\CoraFull}~\cite{bojchevski2017deep} is a well-known citation network that contains labels based on the paper topic. \CoraFull contains a network extracted from the entire citation network of \Cora, while the \Cora dataset is its subset.

\item \textbf{\Physics}~\cite{shchur2018pitfalls} is co-authorship graph based on the Microsoft Academic Graph from the KDD Cup 2016 challenge. Nodes are authors and edges indicate co-authorship. Node features represent paper keywords for each author’s papers, and class labels indicate the most active fields of study for each author.

\item \textbf{\Photo}~\cite{shchur2018pitfalls} is part of the Amazon co-purchase graph, where nodes represent goods, and edges indicate the frequency of two goods being bought together. The node features are bag-of-words encoded product reviews and class labels are given by the product category.
\end{itemize}

\textbf{Implementation Details} We implement both \SFairGNN and the baselines with PyTorch~\cite{paszke2017automatic} and PyTorch Geometric~\cite{fey2019fast}. We initialize model weights with xavier~\cite{glorot2010understanding} and adopt the Adam optimizer to minimize the cross-entropy loss. We set the initial learning rate to $0.005$. We adopt the randomly initialized and trainable embedding vector for each node over a graph instead of node features provided by the dataset. We use $90/10$ as the training/test dataset split.

\subsection{Main Results on Structure Fairness}

To address research question~\textbf{Q1}, we conduct experiments to compare the proposed \SFairGNN with baseline methods and evaluate their fairness performance using the proposed metrics: PCC~(Pearson correlation coefficient) and STD~(standard deviation). In this section, we utilize closeness centrality and eigenvector centrality as indicators of the graph structure in the experiments. Next, we present an analysis of the experimental results for both closeness centrality and eigenvector centrality.

\begin{table}[t]
  \centering
  \caption{Comparison between the proposed \SFairGNN and baseline methods on closeness centrality. The best scores of STD and PCC are printed in bold. \textmd{\textbf{Improv$\uparrow$}} represents the fairness improvement, which is calculated based on the baseline GCN. We run each experiment 5 times and report the means. The hop number is set to $3$. }\label{tab:Accuracy and standard deviation}

\scalebox{0.73}{
    \begin{tabularx}{0.65\textwidth}{X>{\centering\arraybackslash}X>{\raggedleft\arraybackslash}X>{\raggedleft\arraybackslash}X>{\raggedleft\arraybackslash}X>{\raggedleft\arraybackslash}X>{\raggedleft\arraybackslash\columncolor{gray!20}}XXXXX}
    
        \toprule[\heavyrulewidth]
        \multirow{1}{*}{ \textbf{Datasets } } &
        \multirow{1}{*}{ \textbf{Metrics } } &
        \multicolumn{1}{c}{ \textbf{ GCN } } &
        \multicolumn{1}{c}{ \textbf{ GAT } } &
        \multicolumn{1}{c}{ \textbf{ SAGE } } &
        \multicolumn{1}{c}{ \textbf{ \SFairGNN } }  &
        \multicolumn{1}{c}{ \textbf{ $\uparrow$ } } \\
    
     \midrule
     \multirow{3}{*}{ \Cora }    &ACC    &$76.00$            &$66.49$     &$74.85$     &$77.65          $    &$--    $    \\
                                &STD    &$13.15$            &$12.36$     &$12.63$     &$\mathbf{10.53} $    &$19.9\%$    \\
                                &PCC    &$10.47$            &$26.50$     &$10.96$     &$\mathbf{ 0.76} $    &$92.7\%$    \\\midrule
     \multirow{3}{*}{ \CiteSeer }&ACC    &$54.03$            &$47.55$     &$54.45$     &$53.67          $    &$--    $        \\
                                &STD    &$15.35$            &$15.67$     &$16.23$     &$\mathbf{13.07} $    &$14.8\%$    \\
                                &PCC    &$14.50$            &$27.30$     &$14.24$     &$\mathbf{12.92} $    &$10.9\%$    \\\midrule
     \multirow{3}{*}{ \PubMed }  &ACC    &$77.44$            &$66.66$     &$77.82$     &$78.85          $    &$--    $    \\
                                &STD    &$ 8.12$            &$12.48$     &$07.05$     &$\mathbf{5.44}  $    &$33.0\%$    \\
                                &PCC    &$13.48$            &$3.27$     &$14.68$     &$\mathbf{5.98}  $    &$55.6\%$    \\\midrule
     \multirow{3}{*}{ \CoraFull }&ACC    &$49.23$            &$40.53$     &$51.86$     &$50.24          $    &$--    $     \\
                                &STD    &$ 9.65$            &$10.95$     &$09.37$     &$\mathbf{ 7.17} $    &$25.7\%$    \\
                                &PCC    &$24.80$            &$29.35$     &$24.20$     &$\mathbf{15.83} $    &$36.2\%$    \\\midrule
     \multirow{3}{*}{ \Physics } &ACC    &$90.64$            &$78.79$     &$90.61$     &$91.12          $    &$--    $    \\
                                &STD    &$11.33$            &$17.02$     &$ 8.79$     &$\mathbf{ 7.65} $    &$32.5\%$    \\
                                &PCC    &$37.09$            &$50.59$     &$29.95$     &$\mathbf{28.56} $    &$23.0\%$    \\\midrule
     \multirow{3}{*}{ \Photo }   &ACC    &$90.66$            &$89.15$     &$90.81$     &$88.89          $    &$--    $    \\
                                &STD    &$\mathbf{8.71}$    &$10.16$     &$ 8.91$     &$ 8.87          $    &$-1.8\%$    \\
                                &PCC    &$20.03$            &$32.82$     &$15.50$     &$\mathbf{10.19} $    &$48.9\%$    \\
    \bottomrule[\heavyrulewidth] 
    % \end{tabular}
    \end{tabularx}
}
\end{table}

\vspace{5pt}
\noindent\textbf{Structure Fairness with Closeness Centrality}
Here, we show how the proposed \SFairGNN tackles the fairness issue. 
As shown in \cref{tab:Accuracy and standard deviation}, we can see that \SFairGNN achieves the best performance in terms of PCC among all the baseline methods. %As we can see from \cref{tab:Accuracy and standard deviation}, 
The PCC values of \SFairGNN are all close to zero, which indicates that the closeness centrality and classification accuracy are nearly independent. In other words, \SFairGNN successfully tackles the unfairness issue triggered by the biased node egocentric structure. Furthermore, we remark that \SFairGNN has a much smaller STD value than baseline methods on most benchmark datasets. This empirical result indicates that \SFairGNN dramatically improves the fairness of classification decisions for nodes with different closeness centrality scores. Note that the traditional GNNs use an unfair neighbor aggregation strategy to update node representations, causing marginal nodes to receive less information. On the contrary, \SFairGNN expands the neighbors of marginal nodes to debias the unfair node egocentric structures. Such neighborhood expansion strategy enforces structure fairness over the graph. From \cref{tab:Accuracy and standard deviation}, we observe that the overall classification accuracy of \SFairGNN is comparable to the baseline methods. \SFairGNN does not decrease the overall performance even though helping the marginal nodes may harm the nodes at the center of the graph. On the other hand, \SFairGNN can improve structure fairness since the \SFairGNN performs better on both the STD and PCC metrics. Thus our proposed model can alleviate the unfairness issue while maintaining the overall model performance for downstream tasks.

\vspace{5pt}
\noindent\textbf{Structure Fairness with Eigenvector Centrality.} As shown in \cref{tab:pro_pro}, we can see that \SFairGNN achieves the best performance on PCC and STD among the baseline methods.
The PCC values of \SFairGNN are very low, which indicates the eigenvector centrality and classification accuracy are nearly independent. This observation shows \SFairGNN successfully tackles the unfairness issue in the graph. In addition, \SFairGNN has a much smaller STD value than the baseline methods, which means \SFairGNN dramatically improves the structure fairness for nodes with different eigenvector centrality values.

\begin{table}[t]
\centering
\caption{Comparison between the proposed \SFairGNN and baseline methods on eigenvector centrality. The best scores of STD and PCC are printed in bold. \textmd{\textbf{Improv$\uparrow$}} represents the fairness improvement, which is calculated based on the baseline GCN. We run each experiment 5 times and report the means. The hop number is set to $3$.  }\label{tab:pro_pro}
% \vspace{-5pt}
% \begin{tabular}{lcrrrr|r}
\scalebox{0.73}{
    \begin{tabularx}{0.64\textwidth}{X>{\centering\arraybackslash}X>{\raggedleft\arraybackslash}X>{\raggedleft\arraybackslash}X>{\raggedleft\arraybackslash}X>{\raggedleft\arraybackslash}X>{\raggedleft\arraybackslash\columncolor{gray!20}}XXXXX}
        \toprule
        \multirow{1}{*}{ \textbf{Datasets } } &
        \multirow{1}{*}{ \textbf{Metrics } } &
        \multicolumn{1}{c}{ \textbf{ GCN } } &
        \multicolumn{1}{c}{ \textbf{ GAT } } &
        \multicolumn{1}{c}{ \textbf{ SAGE } } &
        \multicolumn{1}{c}{ \textbf{ \SFairGNN } } &
        \multicolumn{1}{c}{ \textbf{ $\uparrow$ } } \\
     
    \midrule
    \multirow{3}{*}{ \Cora }     &ACC    &$76.41$     &$64.97$     &$75.84$                 &$76.00         $        &$--    $       \\
                                &STD    &$ 9.97$     &$12.52$     &$\mathbf{9.96}$         &$11.05         $        &$-10.49\%$         \\
                                &PCC    &$ 9.89$     &$14.20$     &$10.37$                 &$\mathbf{0.03} $        &$99.70\%$      \\\midrule
     \multirow{3}{*}{ \CiteSeer }&ACC    &$53.38$     &$47.61$     &$51.55$                 &$53.42         $        &$--    $       \\
                                &STD    &$16.55$     &$16.18$     &$17.47$                 &$\mathbf{11.85}$        &$28.43\%$       \\
                                &PCC    &$16.60$     &$19.69$     &$14.41$                 &$\mathbf{ 5.96}$        &$64.10\%$       \\\midrule
     \multirow{3}{*}{ \PubMed }  &ACC    &$77.39$     &$65.93$     &$78.44$                 &$77.28         $        &$--    $       \\
                                &STD    &$ 8.24$     &$13.72$     &$ 7.23$                 &$\mathbf{6.03} $        &$26.63\%$      \\
                                &PCC    &$ 4.82$     &$ 6.43$     &$\mathbf{4.53}$         &$5.04          $        &$-4.98\%$     \\\midrule
     \multirow{3}{*}{ \CoraFull }&ACC    &$50.05$     &$40.64$     &$51.90$                 &$47.74        $        &$--    $        \\
                                &STD    &$ 9.21$     &$10.61$     &$ 9.18$                 &$\mathbf{6.94} $        &$24.85\%$      \\
                                &PCC    &$ 9.57$     &$17.34$     &$ 8.99$                 &$\mathbf{5.43} $        &$43.18\%$      \\\midrule
     \multirow{3}{*}{ \Physics } &ACC    &$91.14$     &$76.07$     &$90.63$                 &$91.74         $        &$--    $       \\
                                &STD    &$11.33$     &$18.20$     &$\mathbf{8.23}$         &$9.10          $        &$19.71\%$         \\
                                &PCC    &$17.98$     &$23.34$     &$12.50$                 &$\mathbf{12.41}$        &$44.25\%$       \\\midrule
     \multirow{3}{*}{ \Photo }   &ACC    &$91.02$     &$88.48$     &$91.09$                 &$88.11         $        &$--    $     \\
                                &STD    &$ 8.11$     &$ 9.51$     &$ 8.89$                 &$\mathbf{8.06} $        &$-0.56\%$     \\
                                &PCC    &$ 8.54$     &$ 4.79$     &$ 5.16$                 &$\mathbf{1.64} $        &$80.79\%$     \\
                                
     \bottomrule
     % \end{tabular}
     \end{tabularx}
 }
\end{table}

\subsection{Effect of Fusion Methods}\label{sec:exp:fusion}

To answer the research question~\textbf{Q2}, we conduct a series of experiments to explore the effect of fusion methods in \SFairGNN. We consider replacing the fusion method with several different strategies as follows:
% \begin{itemize}[leftmargin=*]
\begin{itemize}[leftmargin=0.4cm]
    \item \textbf{SEQ} represents the approach where the model aggregates information of different hop neighbors sequentially. This means the model first aggregates information from first hop neighbors and then moves on to aggregate information from other hop neighbors.
    \item \textbf{MAX} represents the approach where the model selects the maximum value of each element from different hop embeddings for each node. This approach highlights the most important neighbor among different neighbors.
    \item \textbf{AVG} represents the approach where the model takes the average of embeddings of different hop neighbors for each node. This approach treats all the neighbors equally during attentive information aggregation.
\end{itemize}

\begin{figure}[!t]
\centering
\subfloat[CiteSeer]{\includegraphics[height=1.2in]{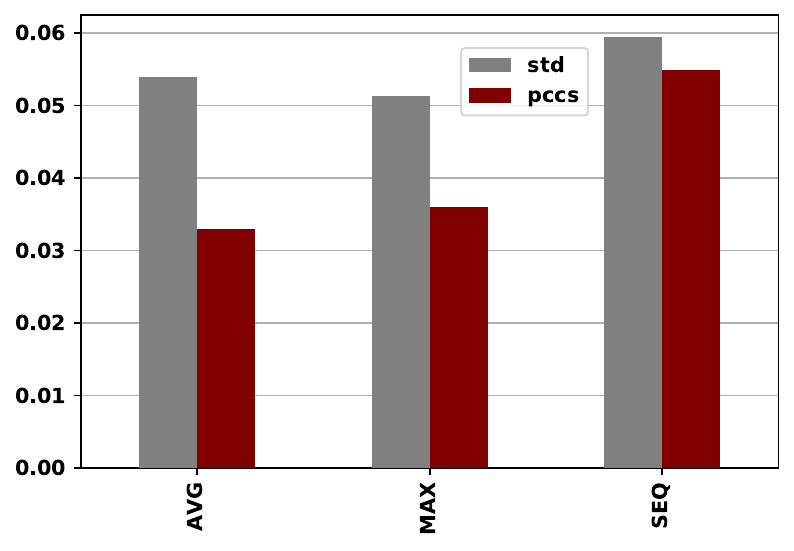}\label{fig:Performance with different fusion methods:CiteSeer}}
\subfloat[CoraFull]{\includegraphics[height=1.2in]{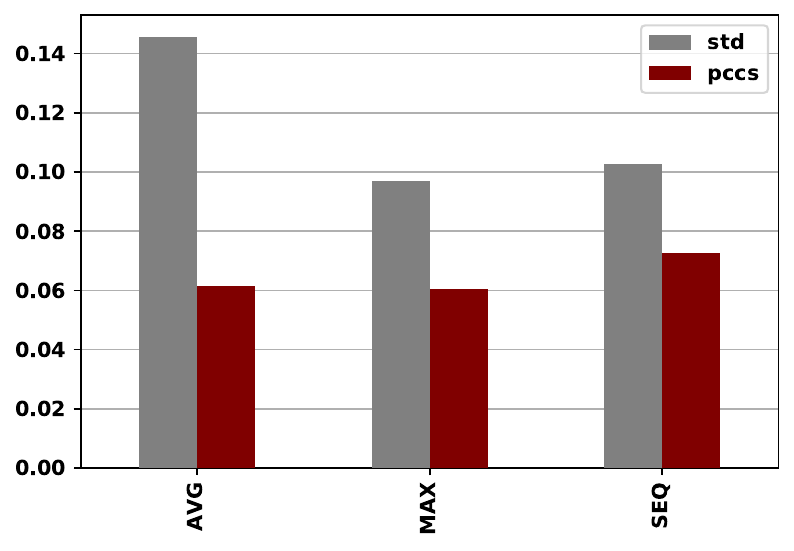}\label{fig:Performance with different fusion methods:CoraFull}}
  \caption{Performance comparison w.r.t. different fusion methods on \CiteSeer and \CoraFull datasets. We use SEQ, MAX, AVG fusion methods. The experiments show that the fusion method MAX obtains the best performance.}
  \label{fig:Performance with different fusion methods}
\end{figure}

One can see from \cref{fig:Performance with different fusion methods} that the fusion method MAX obtains the best performance among the three fusion methods because the MAX fusion method is capable of selecting the most important neighbor automatically to fuse the embeddings. Since the expanded neighbor may introduce noise to the graph structure, the MAX strategy can also reduce the influence of such adverse effects. On the other hand, the SEQ strategy obtains the worst performance on the metric PCC, which may be caused by the fact that SEQ treats different hop neighbors in the same way.

\subsection{Hyperparameter Sensitivity Analysis}
To answer the research question~\textbf{Q3}, we conduct experiments to explore the sensitivity of \SFairGNN to different hyperparameters, including the hop number $h$ and the margin line $line$. The hop number $h$ indicates the extent of the neighborhood expansion. The margin line $line$ determines whether the nodes are marginal nodes or not. For simplicity, we use closeness centrality as the indicator of structure in the experiments for hyperparameter sensitivity analysis.

\begin{figure}[!t]
\centering
\subfloat[CiteSeer]{\includegraphics[height=1.1in]{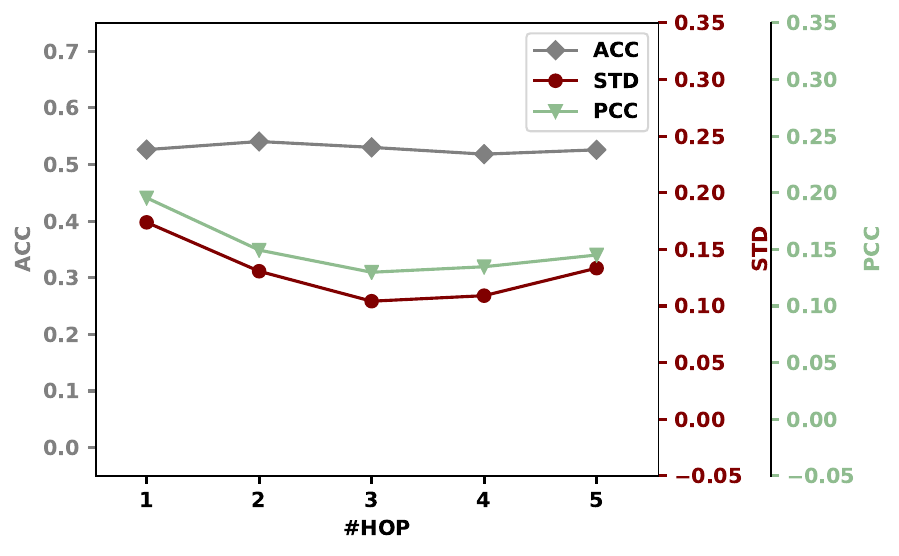}\label{fig:Effect of Hop of Neighbor Coefficient:CiteSeer}}
\subfloat[CoraFull]{\includegraphics[height=1.1in]{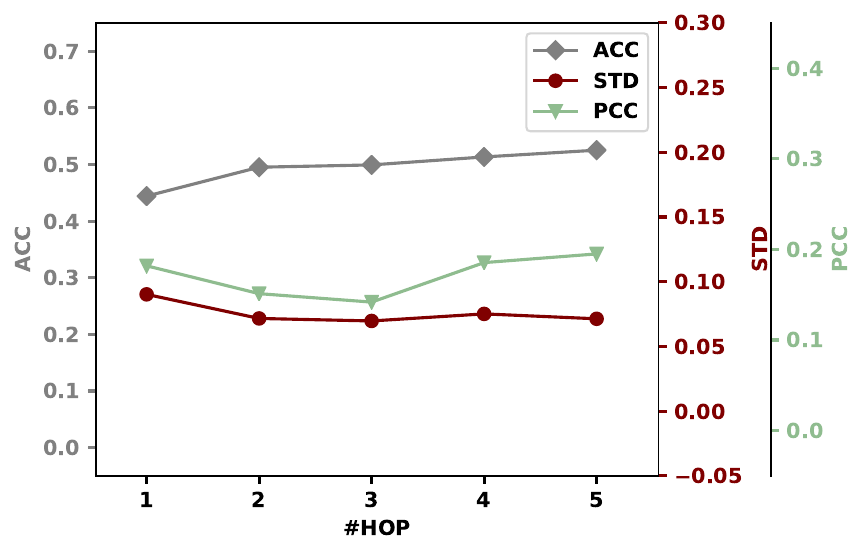}\label{fig:Effect of Hop of Neighbor Coefficient:CoraFull}}
  \caption{The effect of hop number on the \CiteSeer and \CoraFull datasets with the closeness centrality. The results show that 1) the values of PCC and STD generally decrease when the hop number is less than $4$. 2) the performance drops when the hop number is too high.}\label{fig:Effect of Hop of Neighbor Coefficient}
\end{figure}

\vspace{5pt}
\noindent\textbf{Effect of Neighbor Hop.} The hop number $h$ controls the information reception field of marginal nodes, making it a vital hyperparameter. We conduct experiments on the \PubMed and \CoraFull datasets considering a series of hop numbers $h \in \{1, \cdots, 5\}$. We consider closeness centrality for brevity. The results are presented in \cref{fig:Effect of Hop of Neighbor Coefficient}. We observed that 1) the values of PCC and STD generally decrease when the hop number is less than or equal to $4$. This is because the closeness centrality of marginal nodes is improved with the neighborhood expansion, reducing the variation of the nodes' closeness centrality over the graph. 

2) when the hop number is too high, the performance drops because the original graph structure is damaged---being much different from the graph after neighborhood expansion.

\begin{figure}[!t]
\centering
\subfloat[CiteSeer]{\includegraphics[height=1.25in]{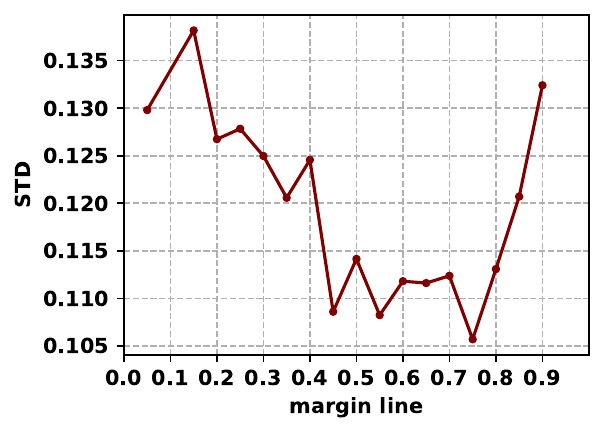}\label{fig:margin line:GCN}}
\subfloat[CoraFull]{\includegraphics[height=1.25in]{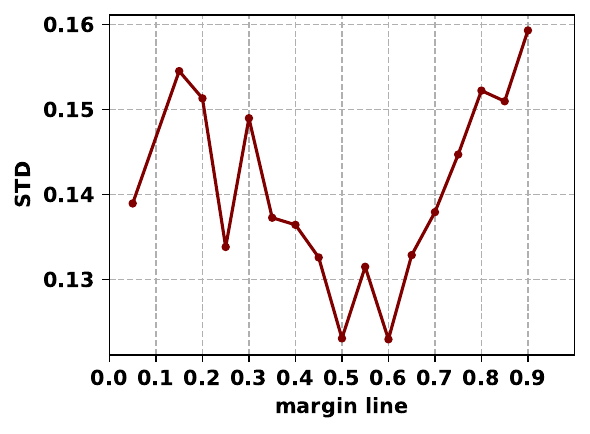}\label{fig:margin line:SFairGNN}}
  \caption{The effect of margin line on the \CiteSeer and \CoraFull  datasets. The results show that the value of STD generally decreases and then increases, where the optimal fairness is reached at the middle of the margin line range.}\label{fig:margin line}
\end{figure}

\vspace{5pt}
\noindent\textbf{Effect of Margin Line.} The margin line $line$ decides whether a node is marginal or not. To explore its effect, we conduct experiments on the \Cora and \CiteSeer datasets with a series of margin line value $line \in \{0.1, \cdots, 0.9\}$ and report the results in \cref{fig:margin line}. One can see from \cref{fig:margin line} that the value of STD generally decreases and then increases, where the optimal fairness is reached at the middle of the margin line range. This is because when the margin line is too small, the marginal nodes obtain insufficient compensation. On the other hand, the neighborhood expansion incorporates too much noise, which hurts the fairness performance when the margin line is too large. In fact, it should be noted that the margin line has significant sociological meanings, so it should be set seriously for specific fairness issues.

\section{Related Works}\label{sec:Related Works}

This section summarizes two categories of related works, including graph neural networks and fairness in graphs.

\subsection{Graph Neural Networks} Graph neural networks (GNNs)~\cite{wu2020comprehensive, zhou2020graph, zhang2020deep,velivckovic2023everything,zhang2022graph,shi2023gigamae,han2022geometric,tan2023bring,zhang2022look} have been the new state-of-the-art method to analyze graph-structured data, which are widely applied to social networks~\cite{huang2019graph}, and academic citation networks~\cite{kipf2016semi}, knowledge graphs~\cite{zhang2020relational,dong2023active}, and to name a few. Starting with the success of graph convolutional network in the semi-supervised node classification task, a wide variety of GNN variants have enhanced and improved the node representation learning and downstream learning tasks~\cite{hamilton2017inductive,velivckovic2017graph,wu2019simplifying}.
Most of GNNs follow a message-passing strategy to learn node representations over a graph. GNNs apply a neighbor aggregator to update node representations iteratively via combining representations of neighbors and that of the node itself. GraphSAGE~\cite{hamilton2017inductive} learns representations by sampling and aggregating neighbor nodes, whereas GAT~\cite{velivckovic2017graph} uses the attention mechanism to aggregate representations from all neighbors. Besides, there is another branch of GNNs, spectral GCNs~\cite{bruna2013spectral, defferrard2016convolutional, kipf2016semi}, which use spectral filters over graph laplacian to define the convolution operation.

\subsection{Fairness in Graphs} Fairness~\cite{mehrabi2019survey,pessach2020algorithmic,caton2020fairness} in machine learning algorithms attracts a lot of interest from both academic and industrial communities since life-changing decision-making needs to be fair in many sensitive environments, such as loan applications, health care~\cite{ahmad2020fairness}, and hiring~\cite{bogen2018help,alder2006achieving}. One mainstream approach to achieve fairness is to use the in-processing approaches, which reduce the bias in the training data. In-processing approaches typically incorporate one or more fairness constraints into the model optimization process to maximize the performance of downstream tasks and improve fairness simultaneously~\cite{dwork2012fairness,zafar2015fairness,zhang2018mitigating}. Recently, a few works~\cite{bose2019compositional,rahman2019fairwalk,dai2020fairgnn,li2021dyadic,ma2021subgroup,jiang2022fmp,dong2022edits,kose2022fair,jiang2022topology} focus on learning fair node representation regarding protected attributes~(e.g., gender, age, race). \cite{bose2019compositional} proposes an adversarial framework to enforce fairness constraints on graph learning, which can flexibly accommodate different combinations of fairness constraints during inference. \cite{rahman2019fairwalk} aims to study potential bias issues inherent within graph embedding and propose a fairness-aware embedding method named Fairwalk. FairGNN~\cite{dai2020fairgnn} uses adversarial debiasing and covariance constraint to regularize the GNN to yield fair predictions. There is also a line of works focusing on the fairness in graph structure~\cite{wang2022uncovering,liu2023generalized,shomer2023toward,chen2022ba}. The position-aware graph neural network~\cite{you2019position} also investigates the ``position'' of the node in graphs, which is related to our proposed method.
A more comprehensive review of related works of fairness in graphs can be found in~\cite{dong2022fairness}.

\section{Conclusion}\label{sec:Conclusion and Future work}
In this work, we propose an \SFairGNN model that is resilient to structure unfairness in graphs. We show that marginal nodes in the graph have lower prediction performance in existing GNN models, which indicates the existing GNNs suffer from the structure fairness issue. To this end, \SFairGNN is proposed to preserve fairness without sacrificing the prediction performance of GNNs. Specifically, \SFairGNN relies on a neighborhood expansion strategy and a hop-aware attentive information aggregation strategy to optimally introduce new information to structurally underprivileged nodes, i.e., marginal nodes. Moreover, we conduct extensive experiments on several graph datasets and validate that our \SFairGNN can achieve fairer node classification performance than the existing GNNs. In the future, we will consider both structure fairness and node attribute fairness for GNNs since the fairness may be caused by the interactions of structure and attributes. Another promising extension of this work is to study the interpretability of structure fairness.

\section{Acknowledgements}\label{sec:Acknowledgements}
We would like to thank all the anonymous reviewers for their valuable suggestions. Na Zou is supported by NSF IIS-1900990, IIS-1939716, IIS-2239257. Fei Wang is supported by NSF 1750326 and 2212175 for this research.

\bibliographystyle{abbrv}
\bibliography{ref}  % sigproc.bib is the name of the Bibliography in this case
\end{document}